\documentclass[letterpaper, 10 pt, conference]{ieeeconf}  

\IEEEoverridecommandlockouts                              

\usepackage{graphics} 
\usepackage[pdftex]{graphicx}
\usepackage{circuitikz}
\usepackage{tikz}
\UseRawInputEncoding
\usetikzlibrary{calc,patterns,decorations.pathmorphing,decorations.markings,decorations.pathreplacing}
\usepackage{tikzscale}
\usetikzlibrary{patterns,decorations.pathmorphing,positioning}
\usetikzlibrary{arrows}
\usepackage{scalefnt}
\usepackage{subfigure}
\graphicspath{ {./images/} }
\usepackage{amsmath}
\usepackage{float}
\usepackage{circuitikz}
\usepackage{pgfplots}
\usepgfplotslibrary{units}
\usepackage{filecontents}
\usepackage{tikz,filecontents,amsmath}
\usepackage{algorithmic}
\usepackage{amssymb}    

\usepackage{bm,times}
\usepackage{bm} 
\usepackage{color} 
\usepackage{cancel}
\usepackage[export]{adjustbox}
\usepackage{mathtools}
\usepackage{caption}
\usepackage{comment}
\usepackage{fixltx2e}
\usetikzlibrary{calc}

\usepackage{multicol}

\usepackage{todonotes}
\usepackage{pdfpages}
\usepackage{graphicx}
\usepackage{caption}
\usepackage{booktabs} 
\usepackage{amsmath,bm,mathtools,upgreek}
\usepackage{amssymb,stmaryrd}
\usepackage{prettyref}
\newrefformat{fig}{Figure~[\ref{#1}]}
\newsavebox\IBoxA \newsavebox\IBoxB \newlength\IHeight
\newcommand\TwoFig[6]{
  \sbox\IBoxA{\includegraphics[width=0.3\textwidth]{#1}}
  \sbox\IBoxB{\includegraphics[width=0.3\textwidth]{#4}}%
  \ifdim\ht\IBoxA>\ht\IBoxB
    \setlength\IHeight{\ht\IBoxB}%
  \else\setlength\IHeight{\ht\IBoxA}\fi
  \begin{figure}[!htb]
  \minipage[t]{0.3\textwidth}\centering
  \includegraphics[height=\IHeight]{#1}
  \caption{#2}\label{#3}
  \endminipage\hfill
  \minipage[t]{0.3\textwidth}\centering
  \includegraphics[height=\IHeight]{#4}
  \caption{#5}\label{#6}
  \endminipage 
  \end{figure}%
}
\makeatletter 
\newcommand*{\b@xplus}[1][+]{\ooalign{%
    $\m@th\vcenter{\hbox{$\m@th#1$}}$\cr%
    \hidewidth$\m@th\boxempty$\hidewidth\cr}} 
\renewcommand*{\boxplus}{\mathbin{\b@xplus}} 
\renewcommand*{\boxminus}{\mathbin{\b@xplus[-]}} 
\makeatother

\usepackage{xcolor} %

\usepackage[colorlinks,citecolor=red,urlcolor=blue,bookmarks=false,hypertexnames=true]{hyperref} 
\hypersetup{linkbordercolor=green}
\usepackage[capitalise]{cleveref}
\usepackage[normalem]{ulem}

\renewcommand{\baselinestretch}{0.98}
\title{\LARGE \bf
RAKOMO: Reachability-Aware K-Order Markov Path Optimization for Quadrupedal Loco-Manipulation
}

\author{Mattia Risiglione$^{1,*}$, Abdelrahman Abdalla$^{1,*}$, Victor Barasuol$^{1}$,\\ Kim Tien Ly$^{2}$, Ioannis Havoutis$^{2}$ and Claudio Semini$^{1}$
\thanks{$^{1}$Dynamic Legged Systems Lab, Istituto Italiano di Tecnologia (IIT), Genova, Italy, \begin{tt}\{name.surname\}@iit.it\end{tt}.
\newline
$^{2}$Oxford Robotics Institute, University of Oxford, UK.
\newline 
$^{*}$Equal contribution.
}
}

\begin{document}

\maketitle
\thispagestyle{empty}
\pagestyle{empty}


\begin{abstract}
Legged manipulators, such as quadrupeds equipped with robotic arms,
require motion planning techniques that account for their complex
kinematic constraints in order to perform manipulation tasks both safely and effectively.
However, trajectory optimization methods often face challenges due to the hybrid dynamics
introduced by contact discontinuities, and tend to neglect leg limitations
during planning for computational reasons. In this work, we propose
\textit{RAKOMO}, a path optimization technique that integrates the strengths
of K-Order Markov Optimization (KOMO) with a kinematically-aware criterion
based on the \textit{reachable region} defined as \textit{reachability margin}.
We leverage a neural-network to predict the margin and optimize it by incorporating it in the standard KOMO formulation. This approach enables rapid convergence
of gradient-based motion planning -- commonly tailored for continuous systems -- while adapting it effectively to legged manipulators, successfully executing
loco-manipulation tasks. We benchmark RAKOMO against a baseline KOMO approach 
through a set of simulations for pick-and-place tasks with the HyQReal
quadruped robot equipped with a Kinova Gen3 robotic arm.
\end{abstract}


\section{INTRODUCTION}

Mobile manipulators are increasingly used in everyday activities due to their
ability to perform tasks in industrial, domestic, and natural environments.
Successful operations in the real world require the robot to reach and manipulate objects while avoiding collisions with both their own structure and the surrounding environment.
The enhanced mobility provided by legs has motivated the robotics community
to explore legged systems as mobile bases for manipulators. For these multi-limbed
systems, referred to as legged manipulators, it is crucial to plan
whole-body motions that are aware of the leg limitations, in order to
avoid compromising the robot's balance and performance in terms of tracking. 
Motion planning specifically addresses this problem by providing
a set of collision-free configurations that respect the robot's limitations. 

In the literature, two powerful techniques for robot motion planning
have been proposed: sampling-based and optimization-based planning,
the latter often referred to as trajectory optimization (TO). Sampling-based methods, such as Rapidly Exploring Random Trees (RRT) and Probabilistic Roadmaps (PRM) \cite{RRT},
can theoretically converge to an optimal solution if one exists, but tend to suffer from a slow
convergence time. By contrast, optimization-based planning
generally yields smoother trajectories that are
compatible with real-time requirements. Despite the risk of converging
to locally optimal or unfeasible robot poses, we adopt
TO in this work given that its convergence time is essential for the advancement of long-term planning \cite{10611389,JPTamp}.
\begin{figure}[t]
    \centering
        \centering
        \includegraphics[width=0.99\linewidth]{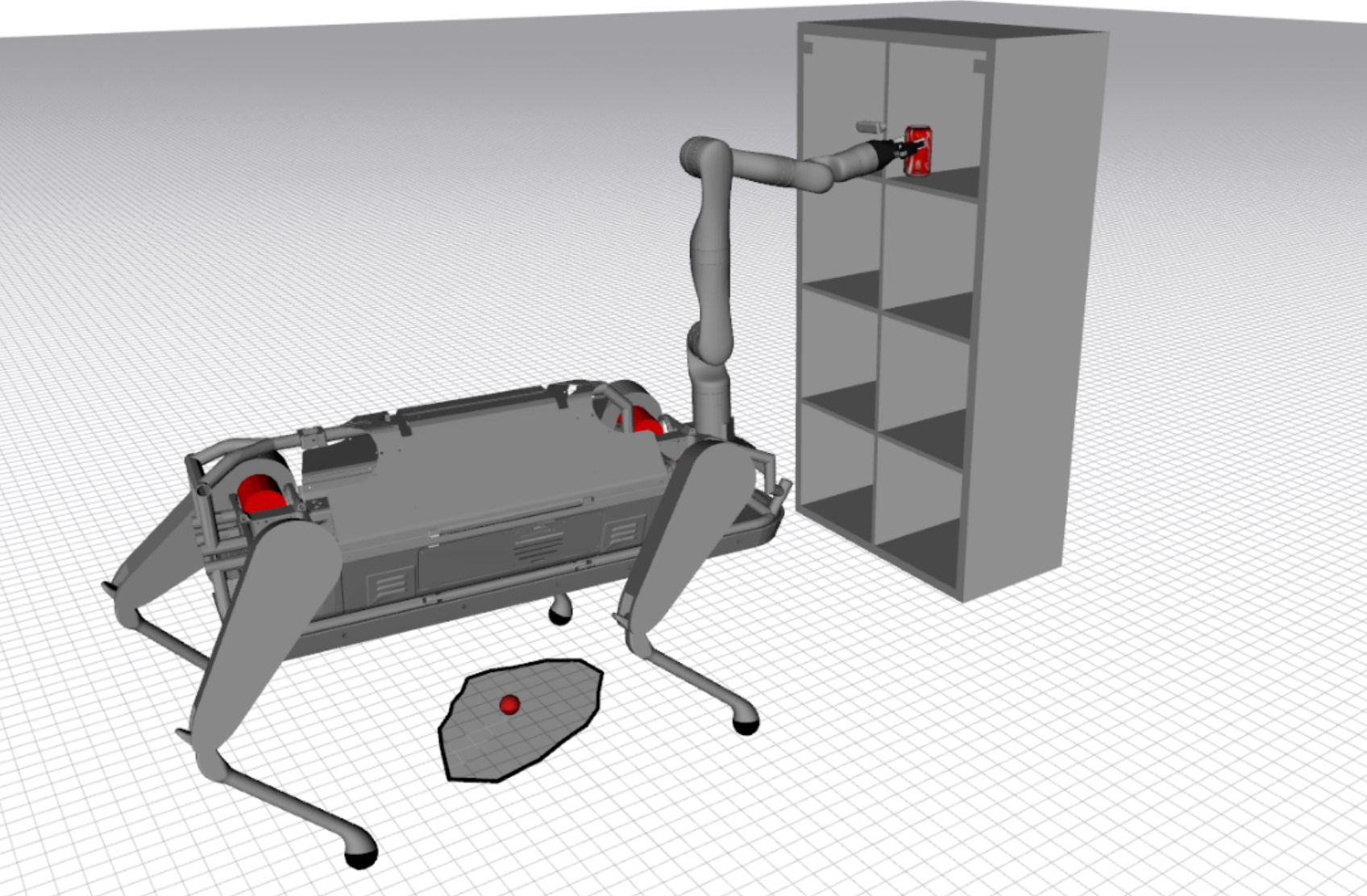}
        \label{fig:figure1}
        \caption{Simulation screenshot of IIT's HyQReal with a Kinova Gen3 robotic arm placing a can on a shelf using a reachability-aware KOMO.}

\end{figure}
Recently, the robotics community has shown significant interest in
using TO to solve task-and-motion planning (TAMP) problems,
where for each task evaluated, a motion planning problem is computed
to test kinematic and/or dynamic feasibility \cite{Toussaint2015LogicGeometricPA}. 
In the general formulation of TO, an objective function prioritizing path length, time, smoothness, and energy consumption is minimized
under running and boundary constraints. Motion planning techniques
such as CHOMP \cite{Chomp}, STOMP \cite{STOMP}, TrajOpt \cite{TrajOpt}
and KOMO \cite{toussaint2014komo} discretize the problem and explore
or approximate gradient information to drive the initial guess to
the optimum.  Their success has been proven for autonomous drone flight, underwater vehicle missions \cite{TrajOptUnderwater},
and pick-and-place tasks with mobile robots (e.g. a PR2 robot \cite{Chomp}).

Trajectory optimization becomes more challenging for legged systems due to their inherently complex dynamical model, driven by nonlinear dynamics, strong coupling among actuated legs and the unactuated base, and the high dimensionality of the overall configuration space.
Additionally, complexity scales up when the
robot's motion has to be optimized for additional limbs that are involved
in non-locomotive tasks. Some prior approaches \cite{Zimmermann2022DifferentiableCA,9561835} address such complexity by simplifying the system to a six
degrees-of-freedom (DOF) base, planning the entire whole-body kinematic motions with
this simplified representation, neglecting the effect of the legs.

The use of reduced-order models \cite{Zimmermann2022DifferentiableCA,9561835}
neglects the leg joint limitations during planning. Different approaches attempt to embed such information in the model without explicitly considering contact forces
nor joint limits.
A simple approach constrains the position of the base relative to the feet through box constraints \cite{box}. These constraints result in conservative approximation of the kinematic limits and provide no metric for optimizing kinematic feasibility.
On the other hand, authors in \cite{proxy}
introduce kinematic feasibility by learning the
probability density of the CoM positions with respect to the
feet locations. Alternatively, researchers also defined
2D regions where if a reference point, e.g., center of mass (CoM) position,
lies inside then: \textit{a)} friction constraints on the contact forces
and motor's actuation limits are respected, namely the \textit{feasible region} \cite{9116813};
\textit{b)} the joint-position limits are respected and leg singularities
are avoided, namely the \textit{reachable region} \cite{10149812}.
A more recent work \cite{orsolino21iros} proposes a function approximator to
estimate the so-called \textit{feasibility margin}, defined as the shortest-distance between the instantaneous capture point (ICP) and the feasible region's boundary. Additionally, the authors in \cite{orsolino21iros}
proposed the margin optimization within a TO formulation to ensure
dynamic locomotion robustness.

\subsection{Contributions}
\begin{figure}[t]
\centering
\includegraphics[width=0.99\linewidth]{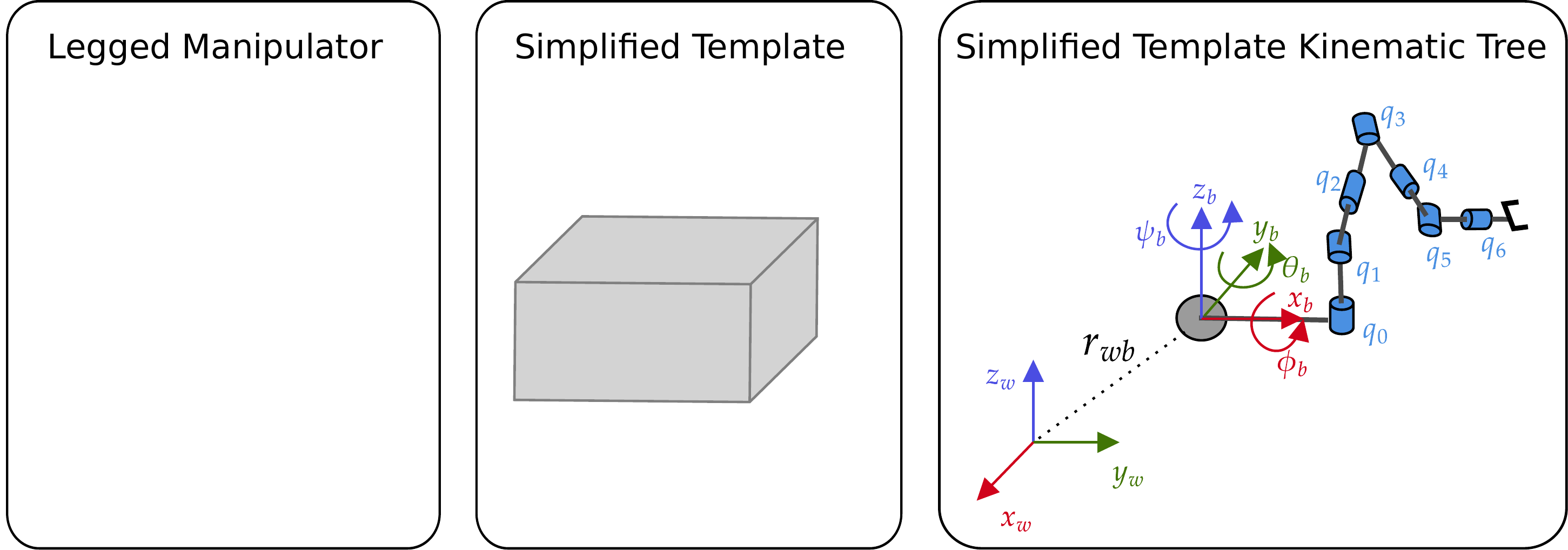}
\caption[Overview of the template model used by KOMO for planning.]{Overview
of the template model used by KOMO for planning. On the left, the full
model of the robot is shown. In the center, the simplified template consists
of a box with a robotic arm attached to it. On the right is the kinematic
tree constructed for KOMO using the simplified template.}
\vspace{-0.35cm}
\label{fig:TemplateRepresentations}
\end{figure}

In this work, we propose an efficient trajectory optimization method for legged loco-manipulation that combines the efficiency of reduced modeling and feasibility of joint-level constraints. 
Explicitly modeling each leg joint introduces $n_c \text{x} n_j \text{x} N$ joint decision variables, where $n_c$ is the number of contacts, $n_j$ is the number of joints in each leg, and $N$ is the planning horizon. This choice significantly increases the problem complexity and computational time. To mitigate this, as proposed in \cite{Zimmermann2022DifferentiableCA, 9561835} we employ a simplified model of the legged robot (such as shown in Fig. \ref{fig:TemplateRepresentations}) at a cost of neglecting any leg joint-level constraint. To address this limitation, we introduce an efficient representation of the joint-kinematic constraints using the the \textit{reachability margin} -- a metric defining the shortest
distance from the horizontal projection of the base position
to the boundaries of the reachable region. 
Inspired by \cite{orsolino21iros}, we utilize a Multi-Layer Perceptron (MLP) to learn the \textit{reachability margin}. We decide to incorporate the network into a KOMO formulation to prevent the robot's base from reaching heights and orientations that would push its legs beyond their reachability limits. This can occur for a legged manipulator whenever the end-effector
has to reach very high or very low targets, as we show in
the results section. An alternative approach could have been to impose empirical limits on the robot's base height and orientations. However, this method risks setting overly conservative constraints. By employing a KOMO formulation, we gain numerical advantages due to the emergence of a banded cost/constraints Jacobian and a banded-symmetric Hessian, as described in \cite{Toussaint2017ATO}. These properties make Gauss-Newton solvers well-suited to solve such problems. The complexity of KOMO is linear on the number of waypoints and polynomial on the dimension of the robot configuration space \cite{toussaint2014komo}. Despite its advantages, KOMO has been applied so far to fixed-based manipulators and omnidirectional wheeled robots \cite{10611389,9981732}.

At each iteration, the KOMO's trajectory is passed to
an inverse-dynamics whole-body (IDWB) controller that solves a one shot
Quadratic Programming problem for generating the robot actuation commands. 
In summary, with the goal of extending
gradient-based planning tailored for non-holonomic mobile manipulators
to legged manipulators and successfully execute these plans online,
we highlight the main contribution as:

\begin{itemize}
    \item We introduce an efficient trajectory planning method for legged manipulators that leverages simplified robot models while effectively incorporating joint-kinematic constraints without resorting to full kinematic representation of the legs. This reduces the number of decision variables by $n_c \text{x} n_j \text{x} N$, resulting only in minimal computational overhead to simplified methods.
   \item We propose a loco-manipulation planning strategy suitable for all legged commercial robots that operate with a black-box locomotion controller, where direct control of individual legs is not available to the control designer \cite{9561835}\cite{Taouil2023QuadrupedalFP}. Instead, the robot can only be steered through velocity inputs.
    The proposed strategy allows to create whole-body motions that inherently account for the kinematic limitations of the legs without explicitly modeling or incorporating them into the planning process, enabling more effective planning motion generation even in the absence of low-level leg control.
 
\end{itemize}

The paper is organized as follows: Sec. \ref{sec:motion_planning} and
Sec. \ref{sec:motion_control} presents, respectively, the motion planning
and control of the robot. Section \ref{sec:results} presents the proposed
simulations and discusses the results. Section \ref{sec:conclusions}
concludes the paper.


\section{MOTION PLANNING}
\label{sec:motion_planning}
In this section, we introduce the \textit{reachability margin} and formulate the reachability-aware formulation of KOMO, named as \textit{RAKOMO}. 
\subsection{Reachability Margin}
In order for the planner to provide kinematically feasible trajectories
when using a simplified template as in Fig. \ref{fig:TemplateRepresentations},
it is crucial to consider the influence of the leg joint limits on the base motion.
However, determining the resulting kinematic limits of the robot base states (e.g. roll, pitch,
and height) is highly challenging because they cannot be treated as independent
from each other. Additionally, such base kinematic limits are strongly dependent
on the relative distance between the base and each foot. Such a relationship, instead,
can be described in an efficient manner using the \textit{reachable region}
introduced in \cite{abdalla20irim}. The \textit{reachable region} describes
the set of all base positions the robot can achieve for the given contact
position and base orientation, where the joint constraints would be respected.
Similar to the approach of \cite{orsolino21iros}, we utilize the
\textit{reachable region} in an optimization problem by defining a metric of
reachability. This metric helps to guide the optimization of the base's motion
to enhance kinematic feasibility and ensure efficient use of the robot's reachable
workspace.

The reachability metric, defined as \textit{reachability margin}, is the
distance between the current base position and the closest boundary of
the reachable region. The computation of the \textit{reachability margin} is achieved
through the following steps: 1) computation of the \textit{reachable region};
and 2) computation of the minimum distance between the horizontal projection of the base position and
the edges of the \textit{reachable region}.

The \textit{reachable region} can be obtained by mapping the kinematic
constraints of the robot (defined in joint space) to the task-space
(the Cartesian space where the base is defined). A ray-casting search is
done to find the furthest point in the Cartesian space where the joint limits
are still respected, i.e., the vertices of the region. As a result from
the nature of the leg kinematic chain, the \textit{reachable region}
is, in general, a non-convex set.

Given that the feasible region is non-convex, we resort to computing the \textit{reachability margin}
by iteratively computing the margins to all the edges of the region (obtained from the vertices) and find the minimum value.
The minimum distance between the base and each edge
$i$ of the region can be computed using
\begin{equation}
    d_i = \frac{|a_i {}_w x_{wb} + b_i {}_w y_{wb} + c_i|}{\sqrt{a^2_i + b^2_i}}
    \label{eq:distance}
\end{equation}
where ${}_w x_{wb}$ and ${}_w y_{wb}$ are the x and y coordinates of the base w.r.t the world frame,
respectively, and $a_i$, $b_i$, and $c_i$ are real constants of the line
equation of the edge, with $a_i$ and $b_i$ being both non-zero. Introducing
$d = \left[d_1, ..., d_{N_e} \right]$, where $N_e$ is the number of edges
of the region, a \textit{reachability margin} $m$ is obtained using $ m = \text{min}\left( \bm{d} \right)$.

To incorporate the \textit{reachability margin} in a gradient-based
optimization problem, we need to be able to compute its gradient w.r.t.
its inputs. Given the non-differentiable nature of the iterative
algorithm used to obtain the region, this can only be achievable
through the finite-difference of the numerical computation. This is considerably
inefficient as the computation of one region can take, on
average, 30 to 40 ms (with the gradient computation requiring over 1 sec). Instead,
we proceed to approximate the \textit{reachability margin} through
training an MLP network.

Note that given it is needed to generate trajectories for the robot's base
(see Section \ref{KOMO}), we have defined the \textit{reachability margin}
in (\ref{eq:distance}) using the robot's base position. However,
in reality, the \textit{reachable region} is only affected by the
configuration of the feet relative to the base. Therefore, to simplify
the training of the network, the input to the \textit{reachability region}
and the MLP can be defined as $   \bm{x_r} = \left[ {}_b \bm{e}_z, {}_b \bm{x}_{{bf}_i}, ..., {}_b \bm{x}_{{bf}_{Nc}} \right]$
where ${}_{b}\bm{e}_z$ is the unit vector representing the z-axis of the world
frame in the robot base frame, ${}_b 
 \bm{x}_{{bf}_i}$ is the i-th foot position
relative to the base frame (and expressed in the base frame), and $Nc$
is the number of feet in contact with the ground. Additionally, to optimize
the margin with respect to the base position, we can use the relationship 
\begin{equation}
{}_{b}\bm{x}_{{bf}_i} = {}_{b}\bm{R}_w ({}_{w}\bm{x}_{{bf}_i} - 
{}_{w}\bm{x}_{wb})
\label{eq:relation}
\end{equation}
where ${}_w \bm{x}_{{bf}_i}$ is the i-th foot position
relative to the base frame (and expressed in the world frame) and ${}_{w}\bm{x}_{wb}$ denotes the base Cartesian position
(expressed in the world frame).
Finally, the network output is the scalar value of the \textit{reachability margin} $m$.

The architecture of the network is composed of 3 hidden layers,
where the first layer has 512 neurons, the second layer has
256 neurons, and the third layer has 128 neurons. The activation
function used in the hidden layers is the ReLU function, while
the output layer has no activation function. The dataset is
generated by sampling the input variables from uniform distributions
centered around nominal operating values. The network
is trained on a dataset composed of $5 \cdot 10^6$ samples, which was verified
to be sufficient to reach a plateau in the prediction performance.
Given that we can warm start the TO within the operating conditions and
that the system is physically constrained to the kinematic boundaries,
generalization outside the selected range is not needed. The time taken
to compute the inference and the gradient of the network
is on average 10 ms and 100 ms, respectively. This allows for the
margin to be utilized efficiently in TO.  

\begin{figure}[h!]
\centering
\includegraphics[width=0.99\linewidth]{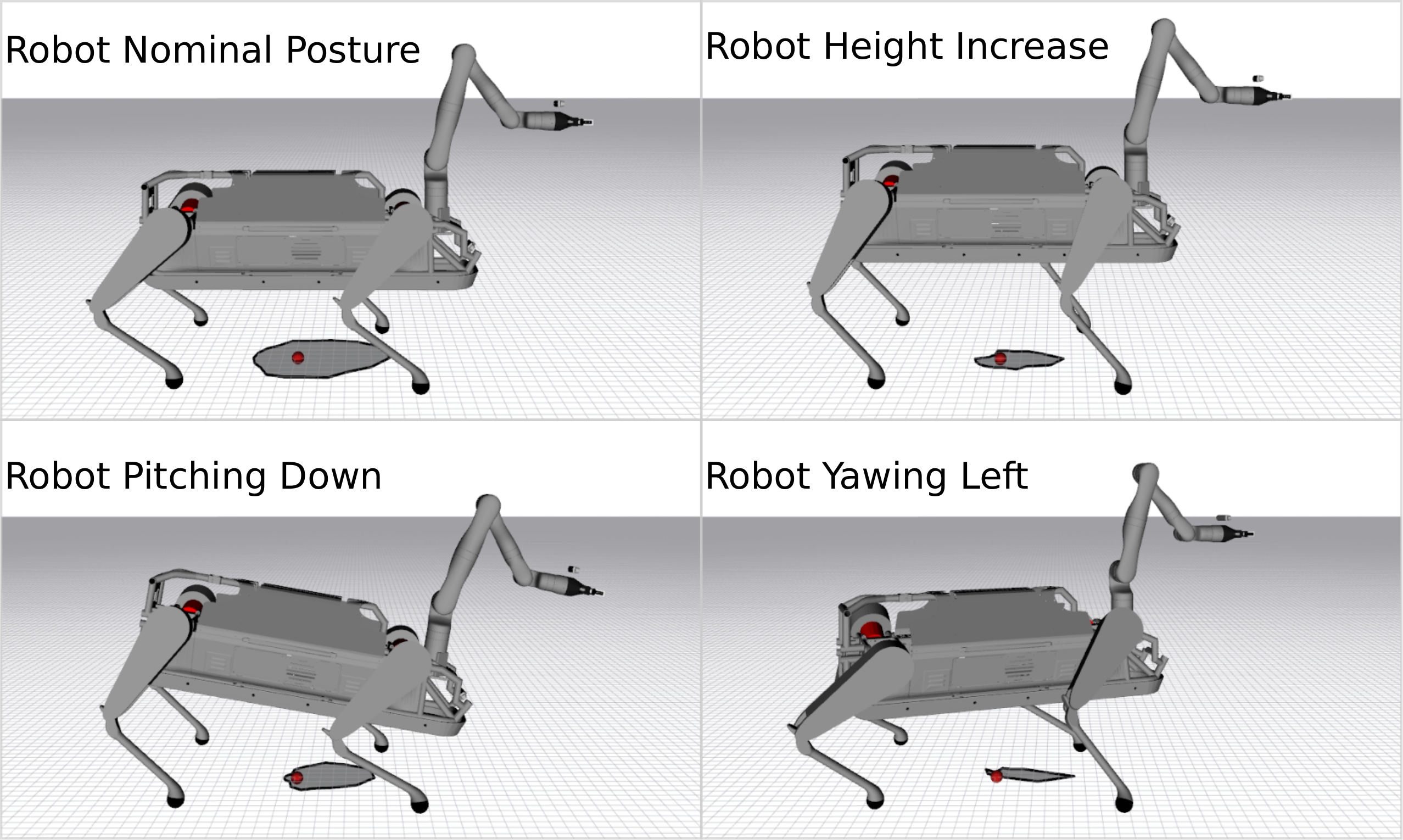}
\caption{Illustration of the reachable region (grey area on the ground) for HyQReal in different postures:
on the top-left, the robot in a nominal posture; top-right, when the robot
height is increased; bottom-left, when the robot pitches down; and
bottom-right, when the robot yaws to the left. The red sphere inside the
region represents the projection of the robot's base position.}
\label{fig:hyqreal_postures}
\vspace{-0.2cm}
\end{figure}

\subsection{K-Order Markov Path Optimization - KOMO}
\label{KOMO}
KOMO is an efficient method for solving motion planning problems
introduced by Marc Toussaint in 2014 \cite{toussaint2014newton}.
It represents a path as a sequence of waypoints, each defined as
a kinematic robot configuration. A robot configuration can be
seen as a set of joints connected by rigid links. We parameterize
the robot configuration space as $\bm q = [ {}_{w}{\bm{x}_{wb}}, {}_{w}{\bm{q}}_{b}, \bm{q}_a]  \in \mathbb{R}^{6+n_a}$ 
where ${}_{w}{\bm{q}}_{b}$ denotes the
base orientation expressed in the world frame using Euler angles,
and $q_1, \ldots, q_{n_a}$ correspond to the manipulator's
joint angles with $n_a$ the number of joints. The goal is to
find a trajectory as a set of robot-configuration ${\bm{q}_0,..,\bm{q}_N}$
that minimizes 
\begin{equation}
\begin{aligned}
\min_{\bm q_{0:N}} \quad &  \bm{f}(\bm{q}_{0:N}) \\
\text{s.t.} \quad & \bm {g}(\bm q_{0:N}) \leq 0, \\
& \bm {h}(\bm q_{0:N}) = 0
\end{aligned}
\label{eq:generalOptFormulation}
\end{equation}
where $\bm{f} \in \mathbb{R}$ is the cost functional,
and $\bm{g} \in \mathbb{R}^{n_g}$ and $\bm{h} \in \mathbb{R}^{n_h}$ are
inequalities and equality constraints, respectively.

The inherent $k$-order Markovian property means that a 
robot configuration at time $t$, defined by $\bm{q}_t$, is influenced by a history of
previous robot configuration up to the $k$-th order, i.e. $\bm{q}_{t-k:t}$.
This property is useful to create costs or constraints only dependent
on a group of $k$-tuplets states, and break older temporal 
dependencies. Hence, we can rewrite \eqref{eq:generalOptFormulation} as 
\begin{equation}
\begin{aligned}
\min_{\bm q_{0:N}} \quad & \sum_{t=0}^{N} \bm{f}_t(\bm{q}_{t-k:t})^T\bm{f}_t(\bm{q}_{t-k:t}) \\
\text{s.t.} \quad  \forall t : \; & \bm {g}_t(\bm q_{t-k:t}) \leq 0, \\
& \bm {h}_t(\bm q_{t-k:t}) = 0
\end{aligned}
\label{eq:komoFormulation}
\end{equation}

with $\bm{f}_t \in \mathbb{R}$, $\bm{g}_t \in \mathbb{R}^{n_{g_t}}$
and $\bm{h}_t \in \mathbb{R}^{n_{h_t}}$. Not all cost and constraint
functions have the same order $k$; it depends on the derivative order
of the kinematic quantity involved. Specifically, for penalizing
positions, velocities, and accelerations, we need respectively $k$
being of order 0, 1, and 2. 

\subsection*{Objective}
The final trajectory should be smooth and penalize overall joint displacement
from the starting robot joint configuration. Hence, we introduce here two type of costs:
\begin{align}
\bm{f}_t &= \|\bm{q}_t - \bm{q}_0 \|^2 \label{eq:PositionBias} \\ 
\bm{f}_t &= \| \bm{q}_t - 2\bm{q}_{t-1} + \bm{q}_{t-2} \|^2
\end{align}

Additionally, we want to optimize the reachability margin $m$ around
a desired value $\epsilon^*$ along the robot's trajectory
\begin{align}
\bm{f}_t = \| \epsilon^* - \textit{m}(\bm {q}_{t})\|^2
\label{eq:MarginOptimization}
\end{align}
where $m$ is a function of the base position and orientation (through
${}_{b}\bm{R}_w$) as shown in (\ref{eq:relation}).

\subsection*{Equality Constraints}
Considering an object to grasp at a certain position (or a target position
to place an object) defined as $\bm z \in \mathbb{R}^3$, we define the kinematic constraint
to reach such desired end-effector's target position as 
\begin{align}
\bm{h}_t = \mathcal{H}(\bm{q}_{t}) - \bm z
\label{eq:Grasping}
\end{align}
being $\mathcal{H}(\bm{q}_{t})$ the differentiable kinematic model of the 
legged manipulator. We adopt a robot-centric perspective, preferring to
express costs for the arm's end-effector which can lead to desired object
motion poses. Hence, we can avoid to increase the dimensionality of the
solution space, by not including rigid objects poses in $\bm q$ during the
different time slices. Handling objects implies including them into the robot's
kinematic tree, once each contact with the object is established. 

\subsection*{Inequality Constraints}
To avoid self-collisions and environmental-collisions in the final trajectory we impose
the following constraints:
\begin{align}
\bm g_t &= -d(\mathcal{K}_i(\bm q_t), \mathcal{K}_j(\bm q_t)) - (r_i + r_j) \le 0 \;\;\; i \neq j \\
\bm g_t &= -d(\mathcal{K}_i(\bm q_t), \mathcal{K}_j(\bm x_o)) - (r_i + r_o) \le 0 \;\;\; \forall i \in \mathcal{S}_r
\label{eq:CollisionConstraints}
\end{align}
with $\forall i,j \in \mathcal{S}_r$, $\forall o \in \mathcal{S}_o$
and $d$ being the Euclidean distance between two collision
primitives $\mathcal{K}$ with sphere-radius $r$ each. The two
sets $S_r$ and $S_o$ denote, respectively, the subsets of robot
and object meshes. Additionally, to respect arm joint limits
and leg joint limits (through the reachability margin) we include
\begin{align}
\bm g_t &= \left\{
  \begin{array}{ll}
   \bm q_{a_t} - \bm{\overline{q}}_a \le 0 \\
   \underline{\bm q}_{a} - \bm q_{a_t} \le 0
  \end{array}
  \right.\\
\bm g_t &= \underline{\epsilon} - m(\bm{q}_t) \le 0 \label{eq:MarginConstraint}
\end{align}
with $\underline{\bm q}_{a}$ and  $\overline{\bm q}_{a}$ being the lower and
upper bound for the arm joint limits, and $\underline{\epsilon} \in \mathbb{R}^{\ge 0}$
a small safety margin threshold. 

\subsection*{Solver}
The formulation of the optimal problem (\ref{eq:komoFormulation}) is general enough for any non-linear programming solver,
given differentiable functions $\bm f_t$, $\bm g_t$ and $\bm h_t$.    
The MLP problem in (\ref{eq:komoFormulation}) is solved with Augmented Lagrangian via Newton's method adopting line search.
The Markovian property leads to having banded Jacobian of the costs/constraints (obtained by partial derivative with respect to $\bm q$) because first-order dependencies are local ($\bm q_t$ depends only on $k$ previous configurations and not all the past states). The non-zero entries of the cost/constraint Jacobian associated to (\ref{eq:MarginOptimization}) and (\ref{eq:MarginConstraint}) are expressed for the rotational coordinates of the base as 
\begin{equation}
 \frac{\partial m}{\partial q_{b}} = (\nabla m_{\bm g})^T \frac{\partial \bm {g}}{\partial q_{b}}   \quad \forall  \, q_{b} \in {}_{w}\bm{q}_b
we\end{equation}
where $(\nabla m_{\bm g})$ is the gradient of the neural network output with respect to the vector $\bm g$. The Jacobian related to the translational coordinates can be obtained as
\begin{equation}
\frac{\partial m}{\partial x_{wb}} = (\nabla m_{{}_b \bm{x}_{bf}})^T \frac{\partial {}_b \bm{x}_{bf}}{\partial x_{{wb}_i}}   \quad \forall \, x_{wb} \in {}_{w}\bm{x}_{wb}
\end{equation}
where $(\nabla m_{{}_b \bm{x}_{bf}})$ is the gradient of the
neural network output with respect to feet positions resulting in $\frac{\partial {}_b \bm{x}_{bf}}{\partial {}_w \bm{x}_{wb}} = - {}_b \bm{R}_w$ from (\ref{eq:relation}).

\subsection*{Collision Avoidance}
In this work, we rely on the Flexible-Collision library (FCL) \cite{6225337} to perform collision and penetration queries. In particular, we leverage the computation of depth penetration between two convex polytopes using the Expanding Polytope Algorithm (EPA) \cite{van2001proximity}, and we use the Gilbert-Johnson-Keerthi (GJK) algorithm \cite{linahan2015geometric} to compute the Euclidian distance between the queried convex shapes used in \eqref{eq:CollisionConstraints}.

\section{Motion Control} \label{sec:motion_control}
The motion planner outputs a set of robot joint configurations, $\bm q$,
containing the base and arm degrees of freedom.\; According to the robot state,
linear interpolation is performed between the current state and the next desired
waypoint from KOMO. The number of waypoints for each dimension of the vector
$\bm q$ is determined as $N = max(\frac{\Delta q_i}{v_i^{max}}f_s) \; \forall i=1,...,6+n_a$,
with $\Delta q_i$ being the displacement between two consecutive configurations
along the i-th dimension, $f_s$ being the control frequency, and $v_i^{max}$
being the maximum velocity of the i-th direction. The desired waypoint, denoted by $\bm q^*$
(containing the base and the arm desired position states), is provided to the dynamic whole-body controller (WBC) \cite{9981895},
with the exception that the desired joint accelerations, denoted by $\bm{\ddot{q}}^d$, are computed as 
\begin{equation}
    \bm{\ddot{q}}^d = \bm K_p (\bm {q}^* - \bm {q}) + \bm K_d (\bm{\dot{q}}^* - \bm{\dot{q}}) + \bm K_i \int_0^t  (\bm {q}^* - \bm {q}) 
\end{equation}
with $\bm K_p$, $\bm K_d$, and $\bm K_i$ being respectively the proportional, derivative, and integral gain matrices. $\bm {q}$ and $\bm{\dot{q}}$ represent, respectively, the current joint position and velocity, while $\bm {q}^*$ and $\bm{\dot{q}}^*$ represent, respectively, the desired joint position and velocity. 

To control the robot torso, a desired wrench $\bm W_{b}^d$ (to be rendered by the WBC) is defined w.r.t. the quadruped's base: 
\begin{align}
\begin{split}
\bm {F}_{b}^d &= \bm {K}_b(\bm {x}_b^d - \bm {x}_b) + \bm {D}_b(\bm {\dot{x}}^d_b - \bm {\dot{x}}_b) 
\end{split} \\
\bm T_{b}^d &= \bm {D}_r(\bm {w}_{b}^d - \bm {w}_b) + \bm {K}_r\bm {e}_r
\end{align}
where $\bm {F}_{b}^d$ and $\bm T_{b}^d$ are respectively the desired
force and moment on the base, i.e ${\bm {W}_{b}^d = [{\bm {F}^d_{b}}^T, {\bm T^d_{b}}^T]}^T$.
We define $\bm {e}_r$ as the base rotational error, $\bm {D}_r\in \mathbb{R}^{3\times3}$
the diagonal derivative gain matrix, and $\bm {K}_r \in \mathbb{R}^{3\times3}$
the diagonal proportional gain matrix. The WBC solves for the stacked vector of
generalized accelerations $\bm {\dot{u}} = [\bm {\ddot{q}}^T_b, \bm {\ddot{q}}^T_l, \bm {\ddot{q}}^T_a]^T \in \mathbb{R}^{6+n_l+n_a}$,
which is composed of the linear and angular accelerations of the base
$\bm {\ddot{q}}_b = [\bm {\ddot{x}}^T_b, \bm {\dot{w}}^T_b]^T \in \mathbb{R}^6$
and the rest of the leg $\bm {\ddot{q}}^T_l$ and arm $\bm {\ddot{q}}^T_a$ joint accelerations.
The WBC optimization problem solves for the vector of decision variables $\bm {\xi} = [\bm {\dot{u}}^T, \bm {F}^T_g]^T$,
from which, using the robot's inverse dynamics, yields desired torques $\bm {\tau} = [\bm {\tau}^T_l, \bm {\tau}^T_a]^T$
computed as:
\begin{equation}
\begin{split}
 \begin{bmatrix}
\bm {\tau}_l\\
\bm {\tau}_a
\end{bmatrix} &=
\begin{bmatrix}
\bm {M}_{lb} & \bm {M}_l & \bm {M}_{la}\\
\bm {M}_{ab} & \bm {M}_{al} & \bm {M}_a
\end{bmatrix}
\begin{bmatrix}
\bm {\ddot{q}}_b \\
\bm {\ddot{q}}_l \\
\bm {\ddot{q}}_a 
\end{bmatrix} +
\begin{bmatrix}
\bm {h}_{l}\\
\bm {h}_{a}
\end{bmatrix} -
\begin{bmatrix}
\bm {J}^T_{st,l}\\
\bm {J}^T_{st,a}
\end{bmatrix}\bm {F}_g \\&-
\begin{bmatrix}
\bm {0}_{lx3n_c}\\
\bm {J}^T_{c,a}
\end{bmatrix}\bm{F}_e
\end{split}
\label{eq:inverseDynamics}
\end{equation}
where $\bm {F}_g \in \mathbb{R}^{3n_c}$ is the robot's foot ground reaction forces, with $n_c$ being the number of contact feet, and $\bm {F}_e \in \mathbb{R}^3$ denotes the external force acting on the arm's end-effector. $\bm {F}_g$ and $\bm {F}_e$ are mapped respectively to the base through the contact Jacobians $\bm {J}^T_{st}$ and $\bm {J}^T_{e}$. $\bm {J}^T_{e,a} \in \mathbb{R}^{6\times n_a}$ is the Jacobian matrix from base to end-effector. 

\begin{figure}[t]
\centering
\includegraphics[width=0.92\linewidth]{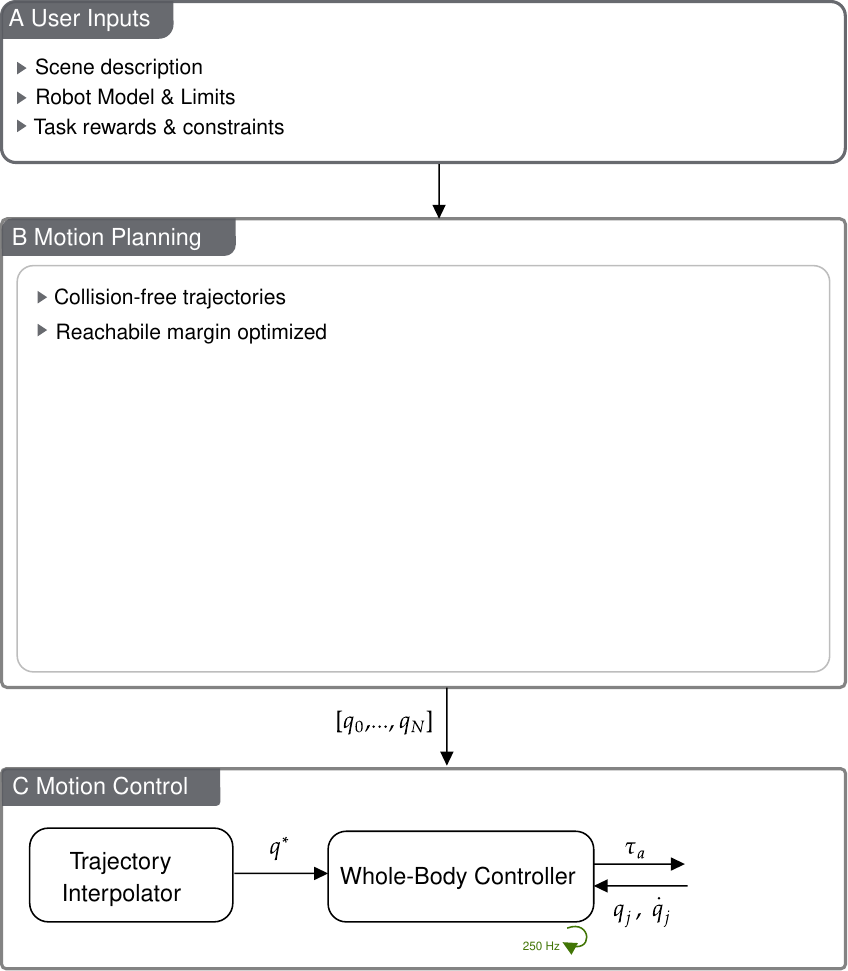}
\caption{Overview of the control architecture for the planning and execution of loco-manipulation tasks. (A) User inputs and task scenario definition. (B) The motion generation part via KOMO calculates collision-free trajectories, taking into account the maximization of the reachable region margin. (C) The trajectories, or desired robot configurations, are interpolated and tracked with a dynamic whole-body controller running at 250 Hz.}
\label{fig:imageOverallTAMPArchitecture}
\vspace{-0.1cm}
\end{figure}


\section{Results}\label{sec:results}
We conducted a set of simulations to validate the proposed approach using our legged manipulator, HyQReal with a Kinova Robotic Gen3 arm.
The simulations were performed using Gazebo as a physics simulator.
Throughout this work we leverage KOMO's solver \cite{ToussaintKOMOTutorial}
to solve full path optimization problems for a fixed number of body configurations $N$ equal to 15. For collision avoidance, we modeled the quadruped's trunk as a bounding box and provided sphere-swept convex meshes around each link of the robotic arm. In all simulations,
we considered the robot in a stance 
 configuration, i.e. with all four feet on the ground and ${}_{w}\bm{x}_{{bf}_i}$ fixed. The integration of dynamic gait and contact sequence optimization is left for future work. 
We designed two simulation scenarios to assess the results of the \textit{baseline}
approach (KOMO without the inclusion of the reachable regions) and our
proposed approach \textit{RAKOMO}: 1) grasping an object from a low height;
and 2) picking and placing an object on high shelves.
All the simulations discussed in this section are also included in the accompanying video.

\begin{figure}[t]
\centering
    \includegraphics[width=0.95\linewidth]{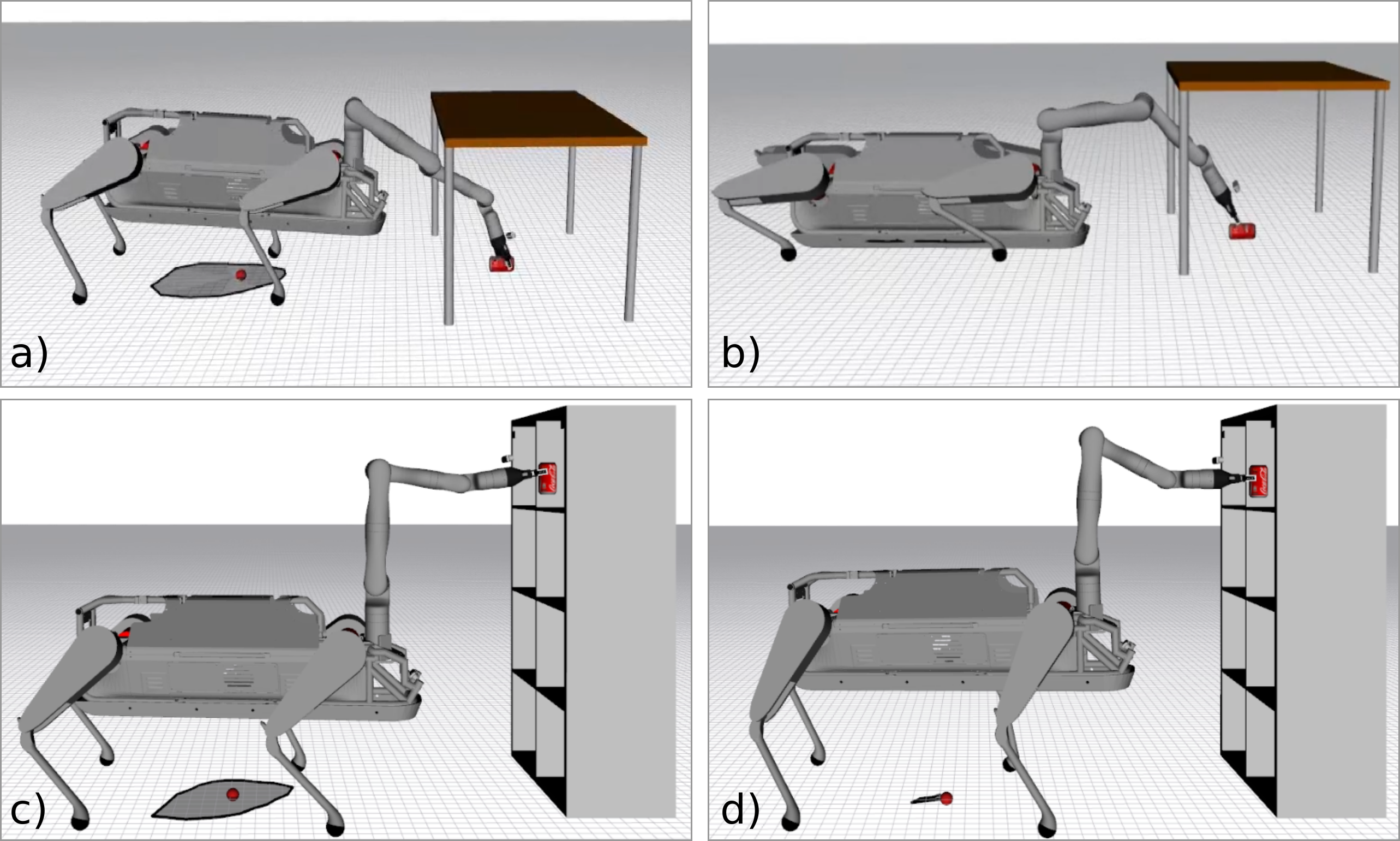}
\caption{Simulation scenarios: grasping low-height object using a) RAKOMO and b) baseline planner;
and picking and placing an object on high shelves using c) RAKOMO and d) baseline planner.}
\label{fig:simulationScenarios}
\vspace{-0.5cm}
\end{figure}

\subsection{Grasping low-height object}
In the first simulation, the robot is required to grasp a bottle underneath a table. The scenario is depicted in Fig. \ref{fig:simulationScenarios}a,b. For the whole task duration, the penalization weight for the arm's position deviations in the cost function \eqref{eq:PositionBias} was set to be 10 times lower than the weight for penalizing base positions. This choice allows to exploit the arm's reachability as much as possible before the optimizer finds body postures in support of the arm. The desired value for the \textit{reachability margin} $\epsilon^*$ in \eqref{eq:MarginOptimization} is set to 0.15$m$ and the minimum feasibility margin $\underline{\epsilon}$ is set to 0.05$m$. 

\begin{figure}[t]
\centering
    \includegraphics[width=0.9\linewidth]{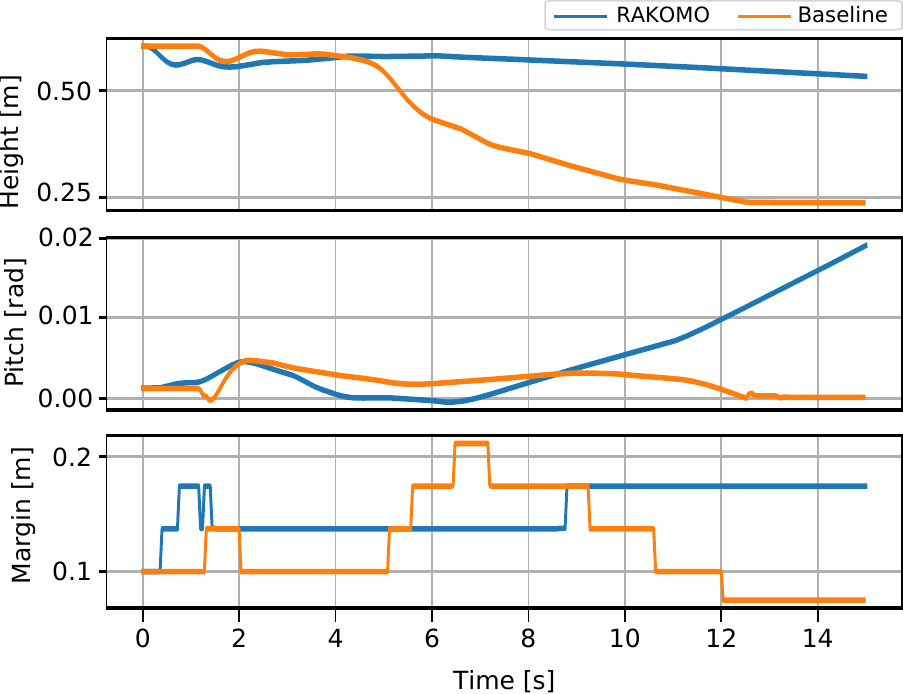}
\caption[Comparison of the \textit{reachability margin} for the executed low-height object grasping motion planned using RAKOMO vs the baseline approach.]{Comparison of the \textit{reachability margin} for the executed low-height object grasping motion planned using RAKOMO and the baseline approach. RAKOMO (blue) achieves a higher minimum value than the baseline approach (orange).}
\label{fig:reachabilityMargin}
\end{figure}

Results showed that introducing the \textit{reachability margin} regularization in  \eqref{eq:MarginOptimization} helps the optimizer to find body poses with better leg kinematic margins along the robot's trajectory. When using the \textit{baseline} approach, the robot is unaware of the leg limitations. As a consequence, the quadruped's trunk lowers excessively, leading to unfavourable leg configurations. This results in a mismatch in the final end-effector position. As shown in Fig. \ref{fig:reachabilityMargin}, the reachability margin starts increasing around 5$s$ because the robot's nominal height starts decreasing, bringing the left-front (LF) Hip-Flexion Extension (HFE) joint towards the center of its kinematic range. However, the same joint retracts more as the base height decreases, resulting in the margin dropping considerably until the motion is stopped when the robot belly hits/touches the ground. On the other hand, \textit{RAKOMO} manages to keep the same joint, i.e. LF HFE closer towards the middle of the workspace throughout the grasping task, resulting in a margin centered around the optimal value $\epsilon^*$.

The possibility of deploying the algorithm on quadruped 
robots, for online replanning, is maintained in both cases:
the time to compute a solution according to the KOMO stopping criteria is 104$ms$
for the baseline and 270$ms$ for RAKOMO.
Although this represents an increase of approximately 2.6 times,
RAKOMO remains significantly more efficient compared to running KOMO
on the full robot model, which would be composed of four additional kinematic chains (legs)
leading to 12 additional joints (see Fig. \ref{fig:TemplateRepresentations}).
Given that the computational complexity scales cubically with the number of joints
\cite{Toussaint2017}, these extra kinematic chains would lead to an estimated computation
time increase of 7 times for the same setup, what would make an online implementation difficult.

\begin{figure}[t]
\centering
    \includegraphics[width=0.9\linewidth]{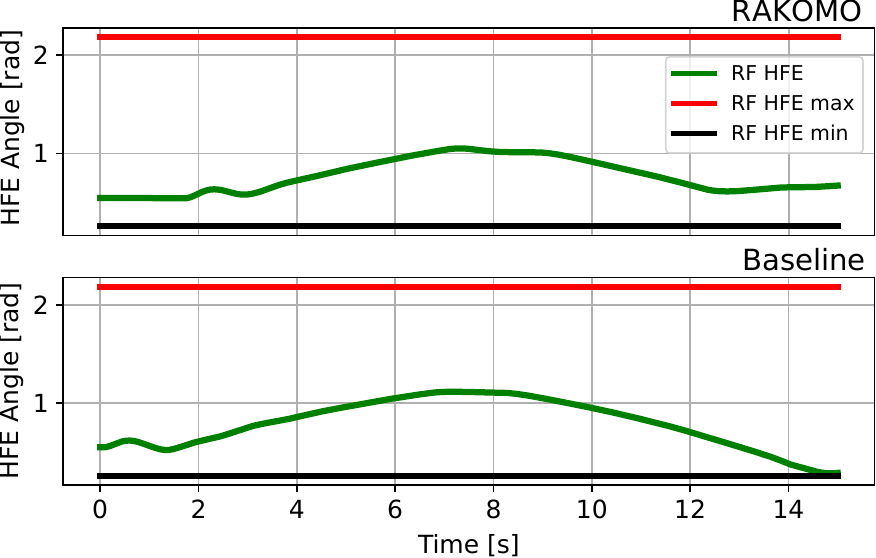}
\caption[Evolution of the HFE joint of the right-front leg during
the execution of the motion.]{Evolution of the HFE joint of the right
front leg during the execution of the high-shelf object manipulation
motion. The red and black horizontal lines represent the maximum and
minimum joint limits, respectively. The optimization of the
\textit{reachability margin} in RAKOMO results in the joint angle
lying closer to the center of the joint range of motion (Top).
Meanwhile, the baseline approach shows the joint angle getting close
to the joint limit (Bottom). }
\label{fig:LFHFEComparison}
\vspace{-0.5cm}
\end{figure}

\subsection{Pick and place of object on a high shelf}
In the second simulation scenario, the robot is positioned in front
of a bookcase and tasked with moving a soda can from a lower shelf
to an upper one. The scenario is depicted in Fig. \ref{fig:simulationScenarios}c,d.
To ensure accurate grasping and placement, we added two 0-order
constraints as shown in (\ref{eq:Grasping}) to enforce the object's
grasping at the initial time and the place position at the final time. 
Both RAKOMO and the baseline approach KOMO explore the planar DOF of
the base to help the arm reach the object and move away from the
shelf before placing the item on the upper shelf. However, during
the final phase of the task, the baseline approach proceeds to raise
the base to an excessive height (as shown in Fig. \ref{fig:simulationScenarios}d)
due to the lack of awareness of the leg kinematic limits. As a result,
the right-front leg (RF) HFE joint reaches its lower bound limit towards
the end of the task, as illustrated in Fig.\ref{fig:LFHFEComparison}.
In contrast, with RAKOMO, the robot's base height and orientation
are maintained closer to more \textit{reachable} values, ensuring
better alignment with the robot's kinematic limits. The difference
in terms of body posture around the end of the task is shown in
Fig. \ref{fig:simulationScenarios}c. From a computational perspective,
the time to compute a solution according to the KOMO stopping
criteria for the solver is 120$ms$, and for RAKOMO it is 285$ms$. 


\section{Conclusions} \label{sec:conclusions}
In this work, we presented a novel methodology for motion planning in legged manipulators that integrates a reachability margin within the K-Order Markov Optimization (KOMO) framework. By leveraging supervised learning to model the reachability margin (i.e. the shortest distance from the base's horizontal projection to the boundary of the leg's reachable region), we incorporated leg kinematic limitations directly into the motion planning process without explicitly modeling all the leg joints. This approach allows for efficient planning on a reduced-order model while ensuring that the generated motions do not push the robot's legs beyond their kinematic limits. We presented simulation results involving IIT's 140kg HyQReal quadruped equipped with a Kinova Gen3 manipulator, performing tasks such as grasping low-height objects and picking and placing items on high shelves. 
The results demonstrated that our method successfully generates whole-body motions that respect leg limitations.
As a future direction, we aim to extend the approach for walking scenarios. Additionally, we aim at considering also dynamic limitations of the robot, such as torque limits.

\bibliographystyle{./bibtex/IEEEtran} 
\bibliography{root}
\addtolength{\textheight}{-12cm}   
\end{document}